\documentclass[10pt]{article}

\usepackage{amsmath}
\usepackage{amssymb}

\usepackage{graphicx}

\usepackage{cite}

\usepackage{color} 

\usepackage[labelfont=bf,labelsep=period,justification=raggedright]{caption}
\usepackage{url}
\makeatletter
\renewcommand{\@biblabel}[1]{\quad#1.}
\makeatother

\date{}

\def\BE{\begin{equation}} \def\EE{\end{equation}}
\def\BEA{\begin{eqnarray}} \def\EEA{\end{eqnarray}}
 \newcommand{\ignore}[1]{}

\begin{document}

% Title must be 150 characters or less
\begin{flushleft}
{\Large
\textbf{What Color is the Sky? (for a non-human) ?}
}
% Insert Author names, affiliations and corresponding author email.
\\
Yair Weiss$^{1}$, 
Ofer Springer$^{2}$
\\
\bf{1} School of Computer Science and Engineering and Edmond and Lili Safra Center for Brain Research, Hebrew Univesity of Jerusalem
\\
\bf{2} School of Computer Science and Engineering,  Hebrew Univesity of Jerusalem
$\ast$ E-mail: yweiss@cs.huji.ac.il
\end{flushleft}

% Please keep the abstract between 250 and 300 words
\section*{Abstract}

The light of the daytime sky contains a mixture of many colors yet is
perceived as blue by human observers. This is largely due to the
particular response functions of the human cones. Under these response
functions skylight and blue light are metamers: they yield the exact
same excitation of the cones.  In this paper we ask: is it possible to
define the ``color'' of the sky for other visual systems?

We present a simple computational method to determine monochromatic
metamers to a given input light for arbitrary visual systems.  Using
published values on spectral sensitivity functions of various species,
we use our method to determine the  dominant wavelength of monochromatic metamers to skylight. For a wide
range of species (bichromats, trichromats and tetrachormats) we find
monochromatic metamers to skylight but the dominant wavelength of the
metamer can vary drastically between species and be very different
from the color perceived by humans.

% Please keep the Author Summary between 150 and 200 words
% Use first person. PLoS ONE authors please skip this step. 
% Author Summary not valid for PLoS ONE submissions.   

% figure 1: rainbow, sky, spectrum and response
% figure 2: CIE diagram
% (end of intro)
% figure 3: Gaussians
% figure 4: species, metamers
% figure 5: summary

\section*{Introduction}
% short explanation of why the sky is blue
% end with CIE diagram: for human cones we can use Helmholtz coordinates.
% question: can we generalize these to arbitrary cone responses? How do we find the %"dominant wavelength" for arbitrary cones? Under what conditions does this %coordinate scheme make sense?

% change the intro to introduce the sky spectrum and the rainbow spectrum

\begin{figure}
\centerline{
\begin{tabular}{cc}
\includegraphics[width=0.5\linewidth]{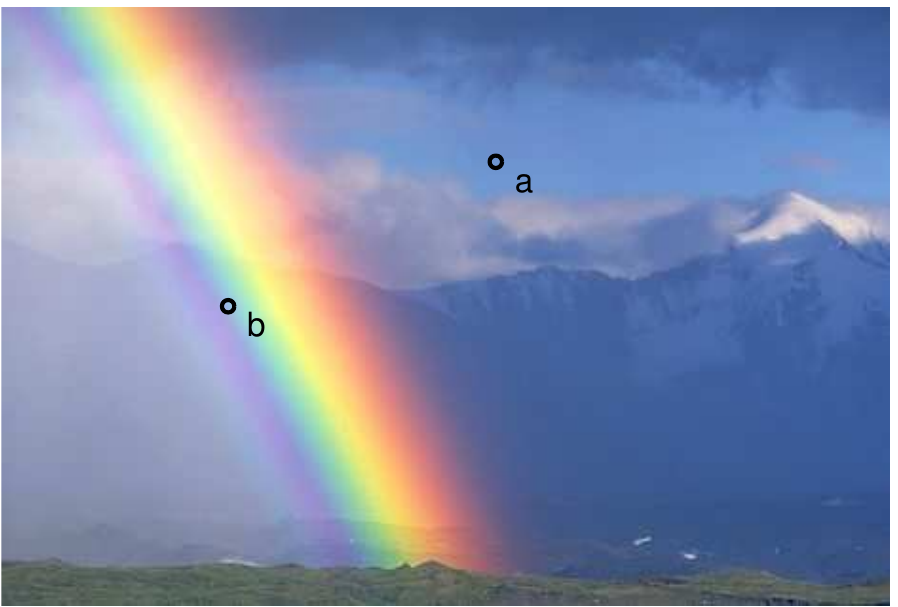} &
\includegraphics[width=0.5\linewidth]{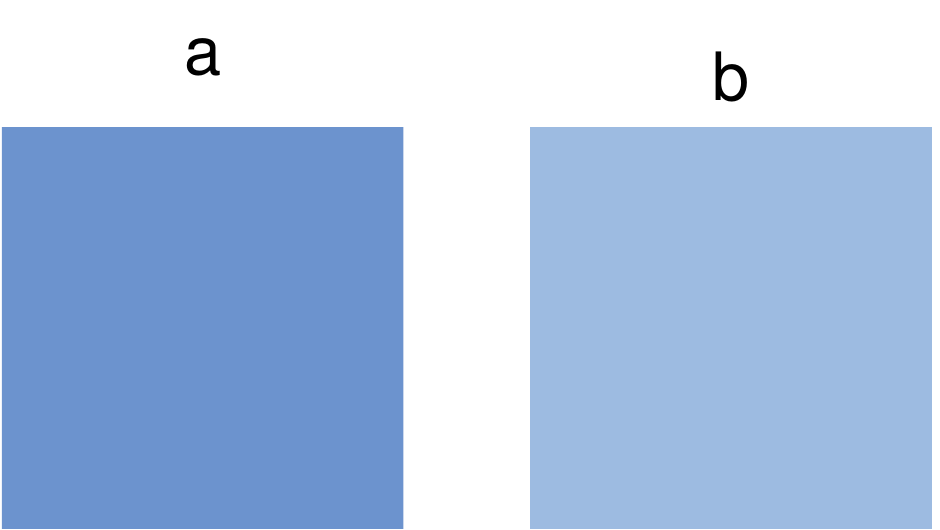} \\
\includegraphics[width=0.5\linewidth]{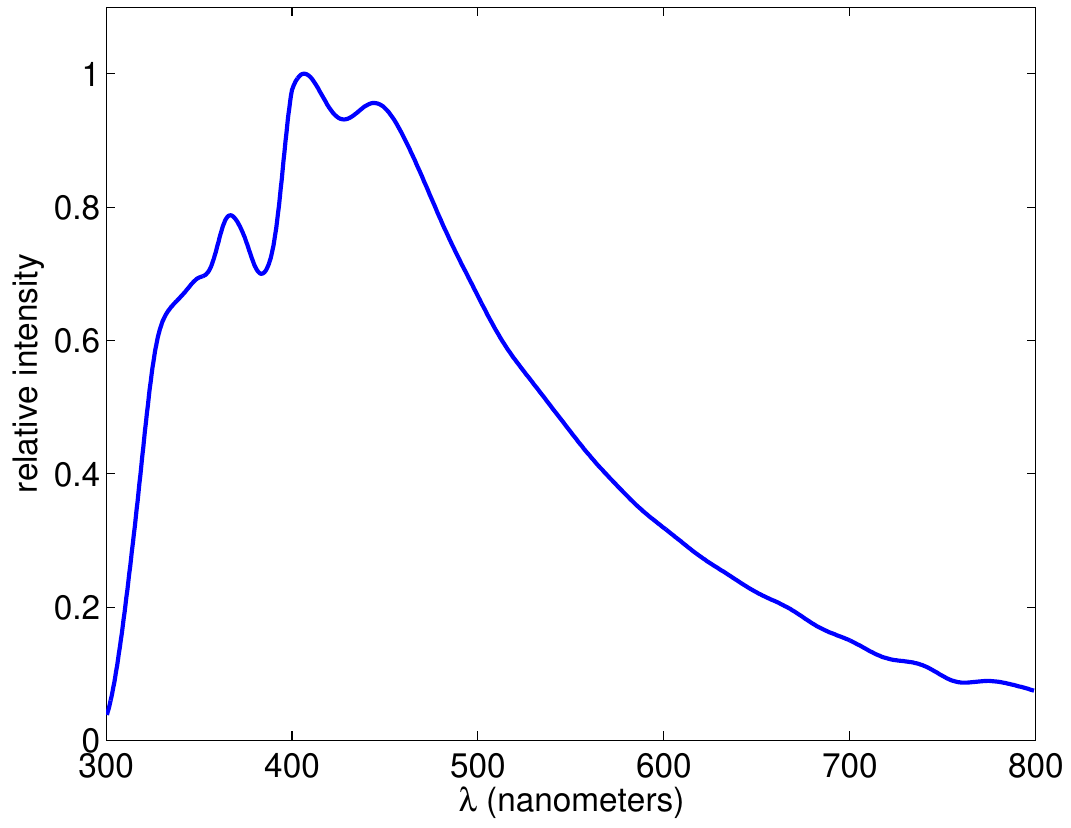} &
\includegraphics[width=0.5\linewidth]{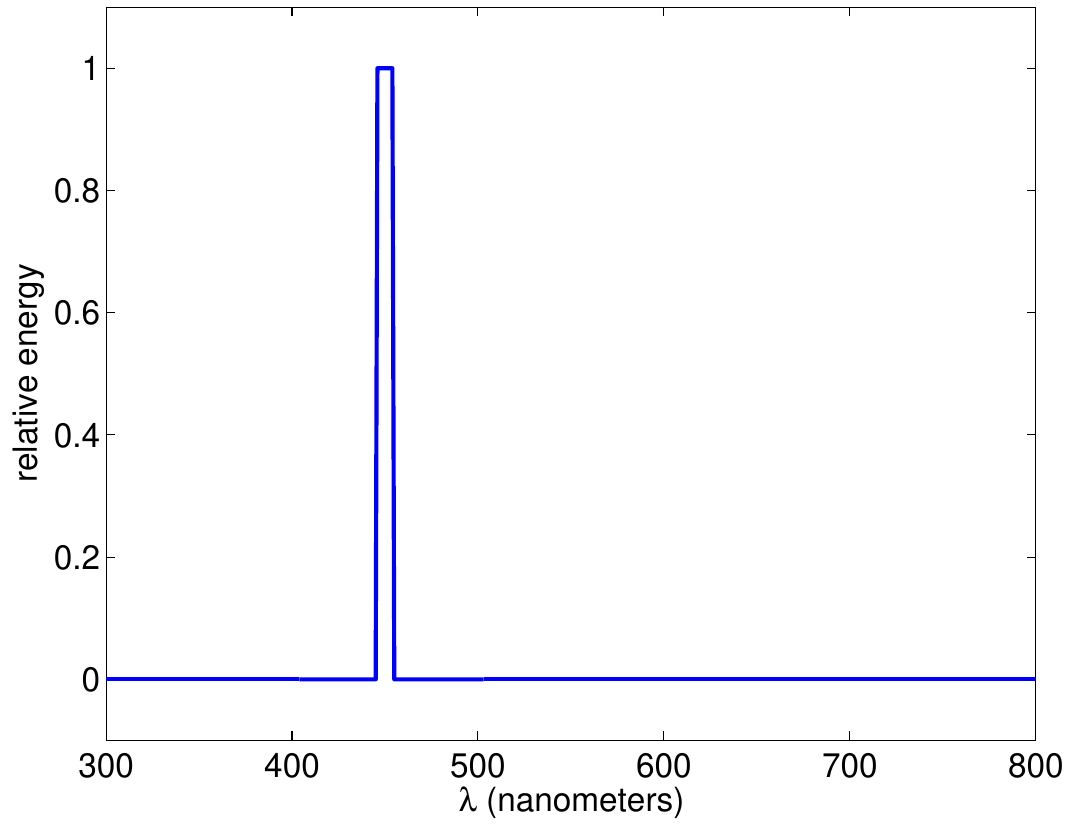} \\
 a & b
\end{tabular}
}
\caption[]{The light from the blue part of the rainbow and the light from the sky are physically very different although we use the word ``blue'' for both.}
\label{intro-fig}
\end{figure}

The light reaching our eyes from the daytime sky is primarily sunlight
that has been scattered by molecules in the atmosphere. While sunlight
is approximately white (containing roughly equal energy at all
wavelengths), the physics of Rayleigh scattering changes the energy
distribution: short wavelength light is more likely to be scattered
than long wavelenth light. As a result, skylight has more energy at
shorter wavlengths compared to longer wavelengths. More precisely, if
$\lambda$ is the wavelength of light then scattering is roughly
proportional to $1/\lambda^4$ and therefore the predicted light of the sky can 
computed by taking the energy
distribution of sunlight and dividing it by $\lambda^4$.
Figure~\ref{intro-fig}a shows this spectrum, which is in excellent
agreement with measurements~\cite{Livingston}.

Perhaps the most salient aspect of the spectral energy distribution of
skylight shown in figure~\ref{intro-fig}a is that it contains a
mixture of all colors. In constrast when we view light that has gone
through a prism or reflected in a rainbow the situation is quite
different (figure~\ref{intro-fig}b). In the ``blue'' part of the
rainbow, the light that is reflected by raindrops is monochromatic: at
any given location only light of a single wavelength is
reflected. Thus the light reaching our eyes from the blue part of the
rainbow has almost all its energy at a single wavelength around
$480nm$.  Why, then, do we use the same word ``blue'' to describe both
lights?

As several authors  have noted~\cite{smith,bohren}, a complete
explanation of the perceived color of the sky must discuss not only
the physics of Rayleigh scattering but also the particular response
properties of human color perception (see also \url{https://xkcd.com/1145/}).  Although skylight contains more
violet light than blue light, it is possible to find a combination of
pure blue light and white light that is {\em metameric} to skylight: that yields the same excitation of the human cones. This is only true for blue light: no combination of violet light and white light will give an exact metamer for skylight. 

Thus there exists a real world scene in which the light reaching our eyes from the second innermost bow of the rainbow and the light reaching our eyes from the sky will give exactly the same excitation to our cones.
But  what
happens if a non-human observer views the same scene? Would the part
of the rainbow that matches the sky be the same one? Will there
necessarily be a match? Will the match be unique?

\begin{figure}
\centerline{
\includegraphics[width=0.5\linewidth]{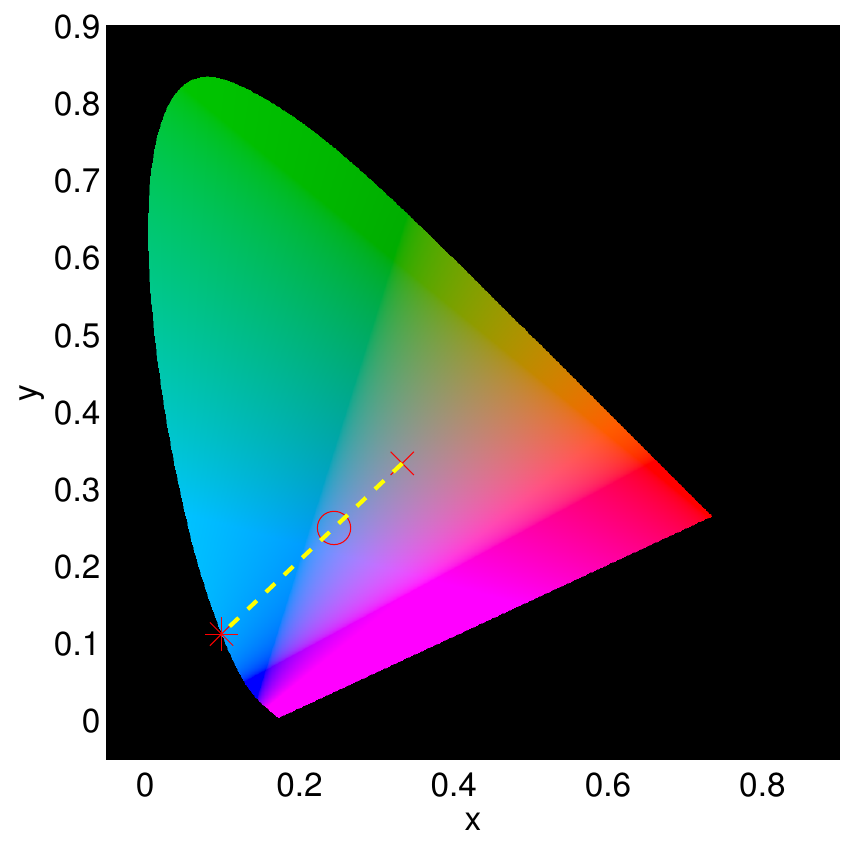}
}
\caption[]{Use of CIE diagram to compute a monochromatic  metamer to skylight. Skylight is denoted by a circle and can be seen to lie on the line between a pure light of approximately 480 nanometer (denoted by an asterisk) and white light (denoted by an x). This diagram is specific to the human cone responses.  In this paper we show how to compute monochromatic metamers for arbitrary visual systems.}
\label{cie-fig}
\end{figure}

The question of how to find a monochromatic light source that is
metameric to an arbitrary light source has a long history in human
color perception and is the basis of what are known as
``Helmholtz coordinates.''~\cite{Stiles}  To transform a light source $l$ into
Helmholtz coordinates one first projects $l$ into CIE chromaticity
space and then passes a line from the white point to the projection of
$l$ (figure~\ref{cie-fig}). The intersection of this line with the curve of
spectral colors determines the ``dominant'' wavelength of the color
$l$.  Since the CIE coordinates are a projective transformation of the physical stimulus, this  also means that $l$ is metameric to a
linear combination of white light and a monochromatic light source
with the dominant wavelength.  For some colors, however, the line from the white point to the light source will not intersect the spectral locus (e.g. for magenta lights) and this means that this light source is not metameric to any part of the rainbow. Thus by plotting skylight in the CIE
chromaticity diagram, we can find the matching part of the rainbow for
a human observer. While there have been different definitions of
chromaticity diagrams for other organisms~\cite{denyaCitation} they
do not allow a simple
translation into human perceptual colors. 

In this paper we show how to define the ``dominant'' wavelength of a
color for an arbitrary visual system. We discuss the conditions under which
any light source can be matched uniquely to a positive combination of
white light and monochromatic light, and present a simple
computational method to determine the wavelength of the monochromatic
light. Applying our method to skylight shows that for many species, a
well-defined ``color'' of the sky exists but the dominant wavelength
can vary anywhere between $400$ and $500$ nm.

\section*{Analysis and Methods}
% analytical results
% the matlab function: find_match

We start by showing how to find a linear combination of white light and monochromatic light that is metameric to a given input light source $l$ for an arbitrary visual system. Let $l(\lambda)$ denote the relative energy of the input $l$ at frequency $\lambda$. We model the visual system as a set of $N$ linear sensors, $c_i(\lambda)$:
\BE
e_i = \int_\lambda l(\lambda) c_i(\lambda) d \lambda
\EE

where $c_i(\lambda)$ are the spectral sensitivity functions of the $i$th receptor. Let $w$ be a white light source whose energy is constant for all wavelengths $w(\lambda)=k$ and $p(\lambda_0)$ be monochromatic light of wavelength $\lambda_0$:
\BE
p(\lambda_0)=\left\{\begin{array}{cc} 1 & |\lambda-\lambda_0|<1 \\
                                     0 & otherwise \end{array} \right.
\EE
Denoting by $Cl$ the mapping of a light source $l$ into the set of $N$ responses $e_i$ we say that two light sources $l,m$ are {\em metameric} if $Cl=Cm$. Thus finding a mixture of monochromatic light and white light that are metameric to an input light $l$ requires finding positive numbers $\alpha,\beta$ such that:
\BE
Cl = C (\alpha w + \beta p_(\lambda_0))
\EE
Using the linearity of the receptors this simplifies to:
\BE
\label{metamer-eq}
Cl = \alpha C w + \beta C p(\lambda_0)
\EE

Equation~\ref{metamer-eq} has a number of linear constraints $N$ which
is equal to the number of color receptors and two unknowns
$\alpha,\beta$. Thus for a trichromatic visual system where $N=3$ this
is an {\em overconstrained} systems of linear equations. There
will either be a unique solution (for the right $\lambda_0$) or there
will be no solutions (for a wrong $\lambda_0$). This simple observation shows that  to find a monochromatic metamer we merely need to exhaustively search over a
range of values of $\lambda_0$ and see whether for that value of
$\lambda_0$ equation~\ref{metamer-eq} has a solution with positive
values of $\alpha,\beta$. 

For visual systems with two color receptors  equation~\ref{metamer-eq}
has two linear constraints and two unknowns. Therefore for many
different values of $\lambda_0$ it will be possible to find a linear
combination of monochromatic light and white light that is metameric
to a given input light. To make the problem well-posed for dichormatic systems, we set $\alpha=0$ and seek a pure monochromatic metamer: we do not allow a mixture of white light and monochromatic light. The equation becomes:
\BE
\label{metamer-eq-di}
Cl = \beta C p(\lambda_0)
\EE
For a fixed $\lambda_0$ this is again an overconstrained problem because it has one 
unknown and two equations. 

% say something about saturation.

When the number of color receptors $N$ is greater than three there
will often not be a linear combination of pure monochromatic light and
white light that is exactly metameric to an input light. For these
settings, we calculate a metamer that is as close as possible to a
linear combination of white light and monochromatic light. 
Formally, we seek a metamer $m$ such that $m=\alpha w + \beta
p_(\lambda_0) + n$ 
where $n$ is a perturbation vector that ideally would be zero. 
We search for $(\alpha,\beta,\lambda_0)$ such that the norm of $n$ is
minimal. For a fixed $\lambda_0$ it can be shown that $\alpha$ and
$\beta$ can be found by solving:
\BE
\left( M^T Q M \right) \left( \begin{array}{c}
                        \alpha \\ \beta
     \end{array} \right)
= M^T Q C l
\EE
where $M$ is a matrix whose two columns are $w$ (white light) and
$p(\lambda_0)$ (pure monochromatic light at wavelength
$\lambda_0$) and the matrix $Q$ is given by $Q=\left(C C^T\right)^{-1}$. 

Given $\alpha$ and $\beta$ we can then solve for $n$ which is given by
\BE
\label{n-equation}
n = C^T Q \left( C l - C M \left( \begin{array}{c}
                        \alpha \\ \beta
     \end{array} \right) \right)
\EE
Our algorithm therefore searches for a $\lambda_0$ for which the norm
of $n$ in equation~\ref{n-equation} is minimal. 

% two remarks: (1) does not depend on rescaling of $c_i$ and (2) uniqueness. 

% Results and Discussion can be combined.
\section*{Results}

\begin{figure}
\begin{tabular}{c}
Cones:\\
\includegraphics[width=0.3\linewidth]{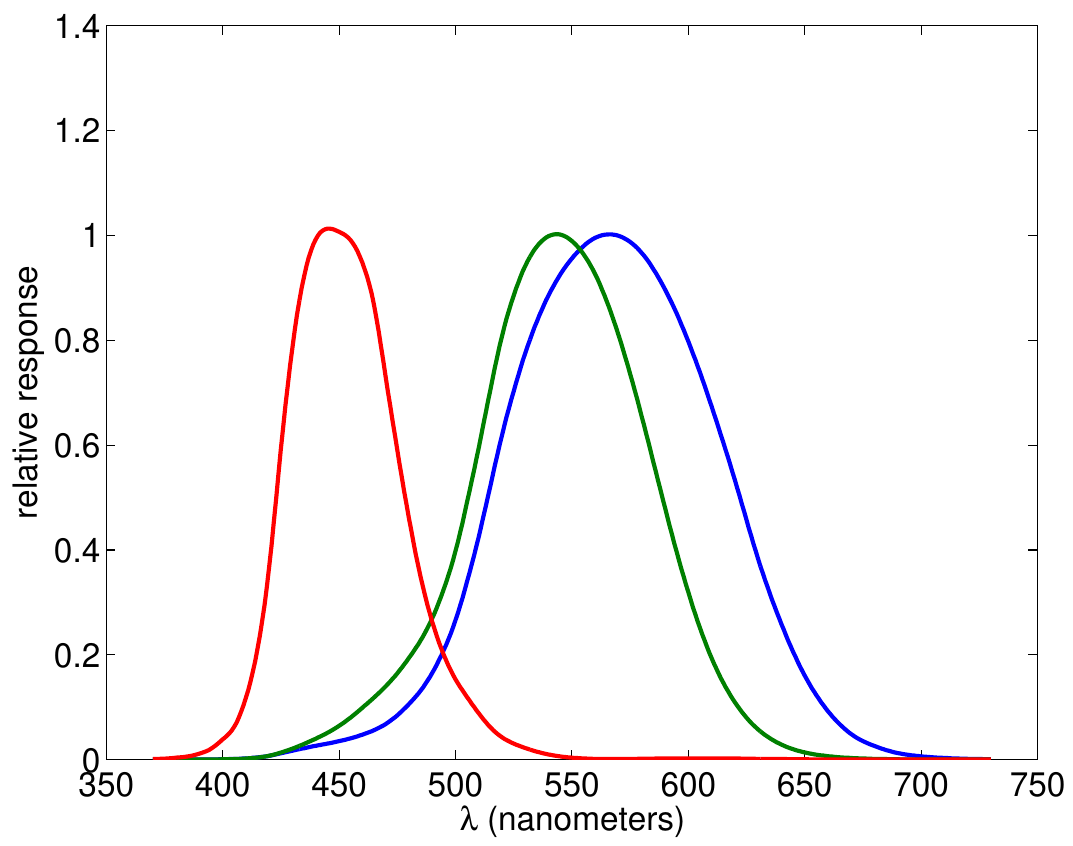} 
\end{tabular}
\\
\begin{tabular}{cc}
Physical Input & Cone activation \\
\includegraphics[width=0.3\linewidth]{Figs/skyspectrumlowpass.pdf} &
\includegraphics[width=0.3\linewidth]{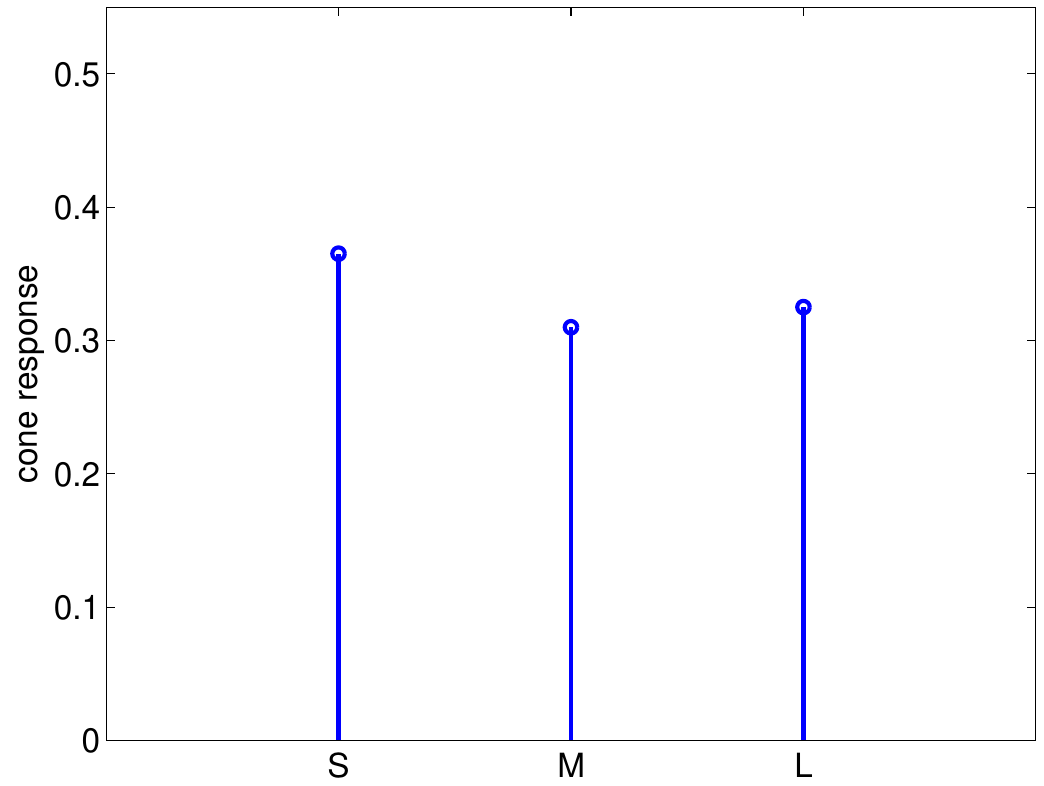} 
 \\
\includegraphics[width=0.3\linewidth]{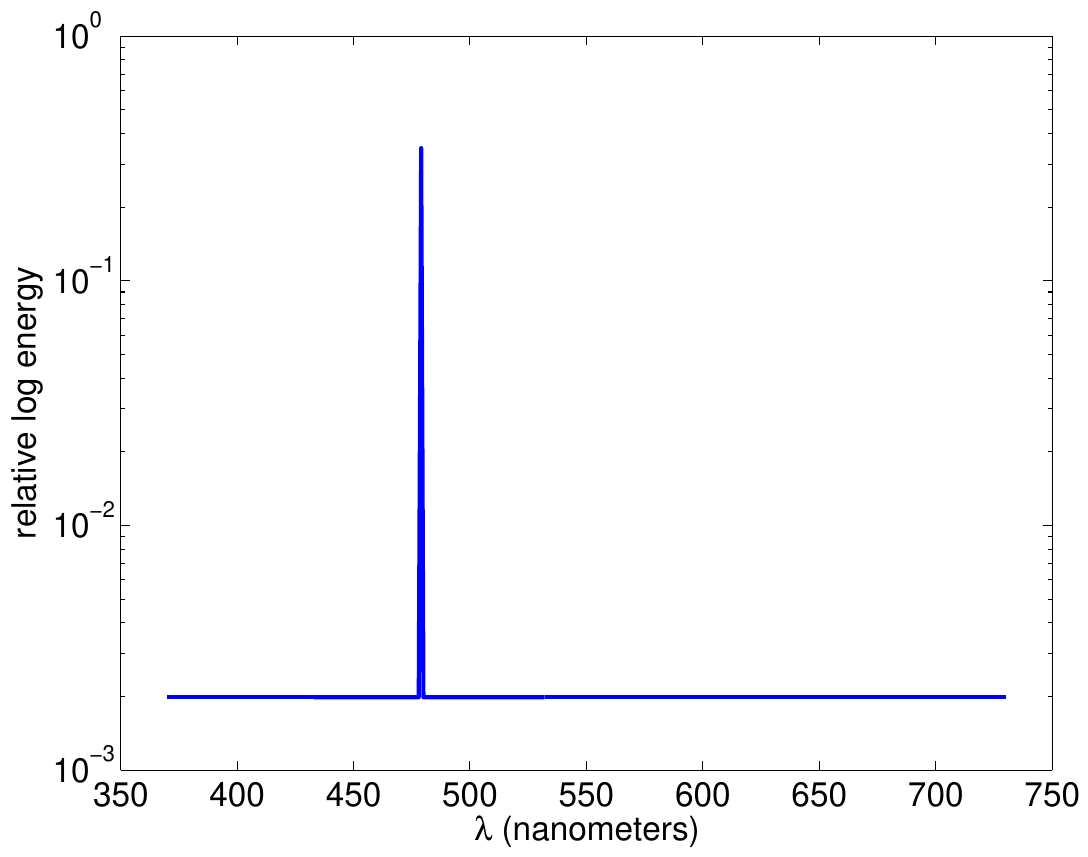} &
\includegraphics[width=0.3\linewidth]{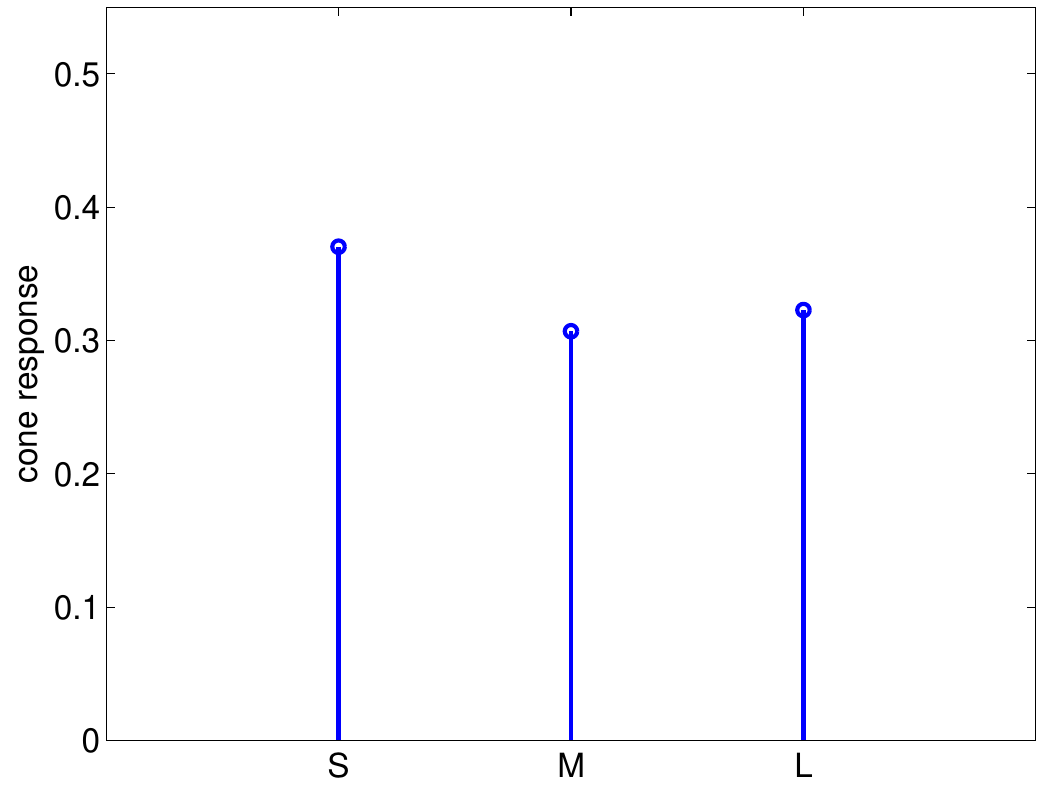}  
\end{tabular}
\caption[]{Top: the human cone responses. Bottom: skylight and the monochromatic metamer found by our algorithm  that gives exactly the same cone response as skylight. For the human cones, the wavelength of the monochromatic light is $479nm$.}
\label{human-metamer}
\end{figure}
 
In our first experiment we used our algorithm to replicate  the standard result that for the human visual system the monochromatic metamer to skylight is approximately $480nm$~\cite{smith,bohren}.  We set the input light $l$ to be the sky spectrum shown in figure~\ref{intro-fig}a and the cone matrix $C$ to be the human cone responses~\cite{wandell1995foundations}. We then used the algorithm described in the previous section to find a mixture of pure monochromatic light and white light that is metameric to the input $l$. The bottom of  figure~\ref{human-metamer} shows our result: we show the computed metamer as well as the computed cone responses for both skylight and the metamer.  As expected, the metamer is around $480nm$ (specifically it is at $479$nm).

In displaying the cone responses in these figures we use an arbitrary scaling of the responses so that all cone response functions have a maximum of one. However none of our results will change if the response functions are rescaled. This is because we are dealing with {\em exact} metamers, so if two light sources give exactly the same response under one rescaling this will still be true under another rescaling.

\begin{figure}
\begin{tabular}{cccc}
\includegraphics[width=0.2\linewidth]{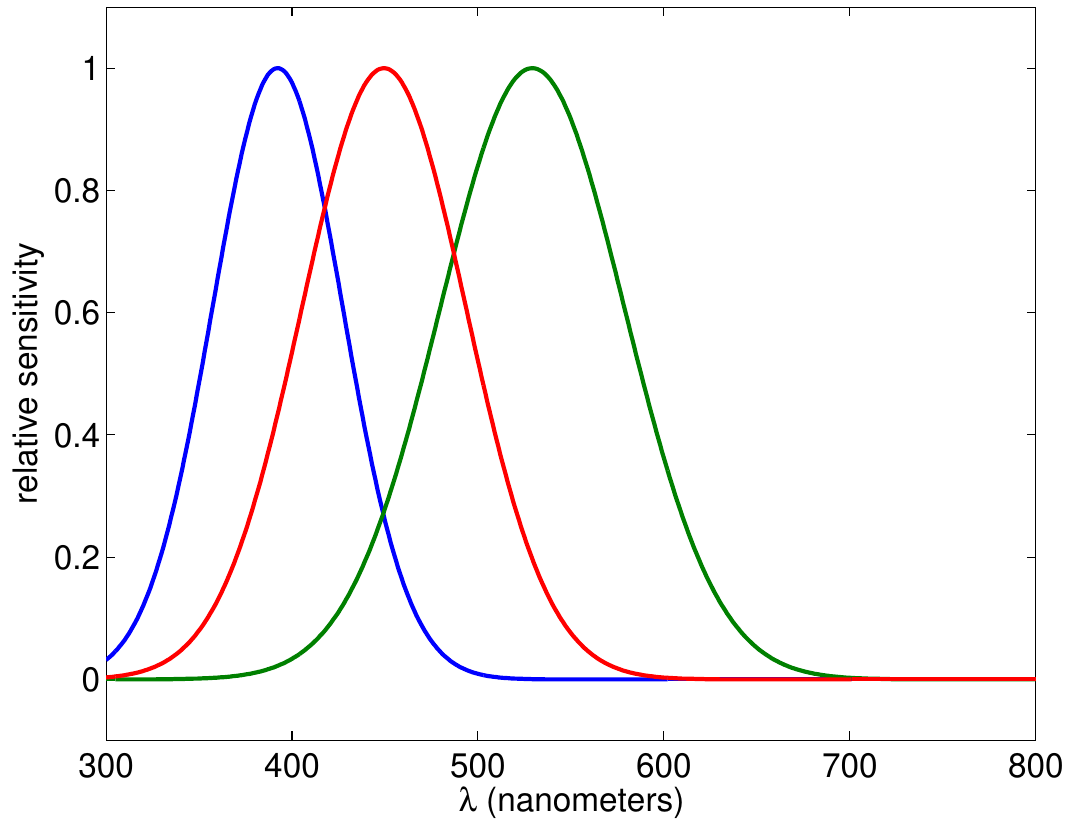} &
\includegraphics[width=0.2\linewidth]{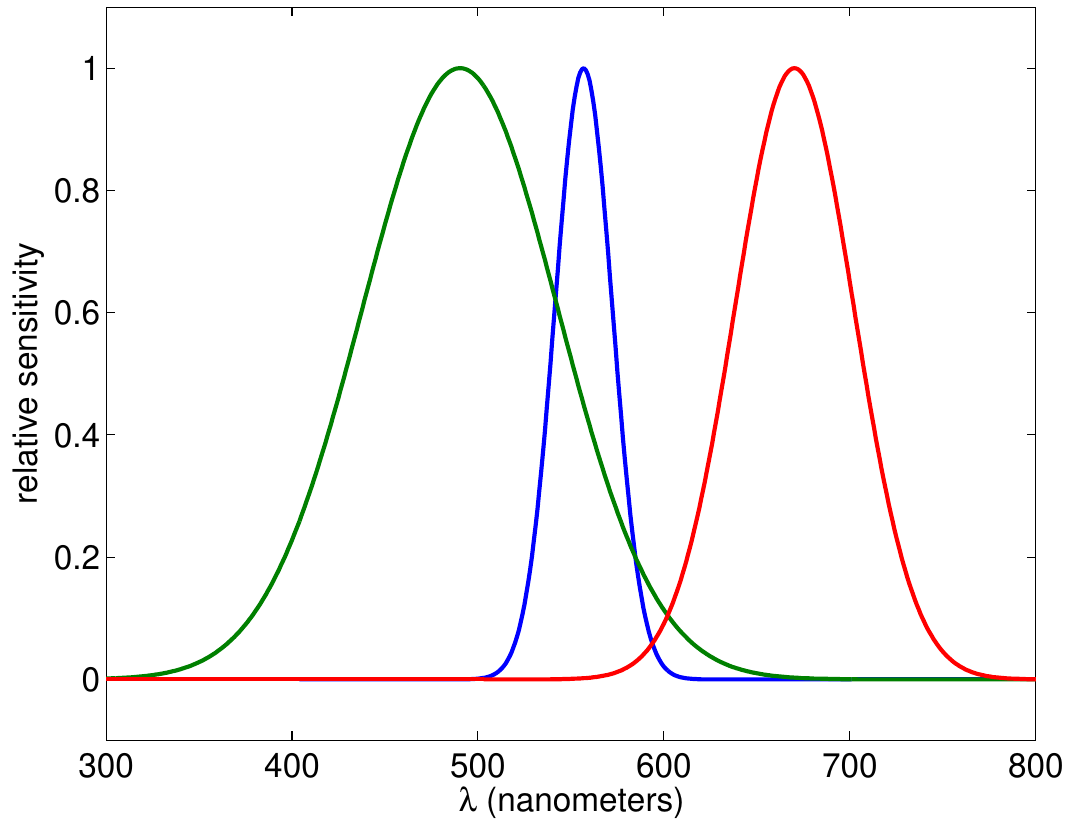} &
\includegraphics[width=0.2\linewidth]{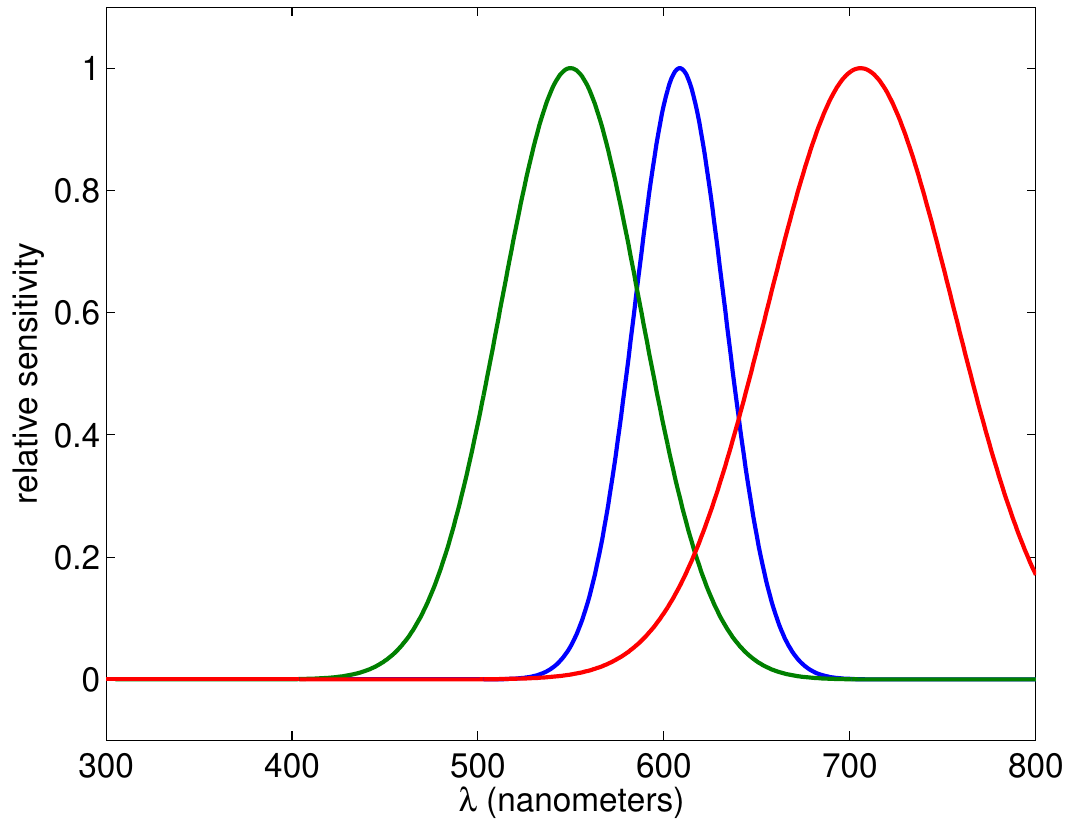} &
\includegraphics[width=0.4\linewidth]{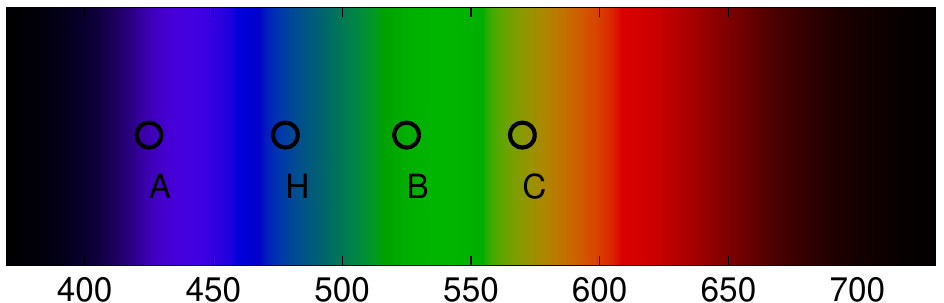} \\
A ``cones'' & B ``cones'' & C ``cones''  & D: Perceived sky color
\end{tabular}
\caption[]{ {\bf A-C} Three synthetic visual systems generated randomly. {\bf D:} The wavelength of the monochromatic metamer to skylight can vary between $400$ ('violet') and $580$nm ('yellow').}
\label{synthetic-fig}
\end{figure}

In a second set of  experiments we wanted to determine the extent to which the perceived color of the sky is dependent on the cone response properties. We randomly generated synthetic trichromatic visual systems whose cone responses are Gaussian functions. The peak response of each cone was chosen randomly in the visible spectrum (between $400$ and $700$nm) and the width of the Gaussians was chosen randomly (standard deviation  between $10$ and $60$nm). For each such synthetic cone matrix we ran our algorithm and determined the wavelength of the monochromatic metamer. Figure~\ref{synthetic-fig} show three selected randomly created cone matrices. For these three aritifical visual systems  the exact same physical stimulus yield metamers with wavelengths of approximately $400$nm ('violet'), 540nm ('green') and 580nm ('yellow').  Figure~\ref{synthetic-fig}d plots the dominant wavelengths of the metamer to skylight for human cones and the three synthetic cones. Clearly, the physics by themselves do not determine the perceived color.

In a third set of experiments we use published values of spectral sensitivity functions of different species and calculate the monochormatic light that would be metameric to skylight for each species. We simulated the spectral sensitivity function using a standard visual pigment template~\cite{govardovskii2000search} :
\BE
c(\lambda;\lambda_{max})=\frac{1}{e^{A(a - \lambda_{\max}/\lambda)} +
e^{B(b - \lambda_{\max}/\lambda)} + e^{C(c - \lambda_{\max}/\lambda)}  +D}
\EE
with $A=69.7$,$B=28$,$b=0.922$,$C=-14.9$,$c=1.104$, $D=0.674$ and 
 $a=0.8795+0.0459\exp(-(\lambda_{\max}-300)^2/11940)$.

% human, dog, gerbil, gecko, honeybee,  chicken, butterfly 
\begin{table}
\begin{tabular}{|c|c|c|c|}
\hline
Species & Cones & metameric $\lambda^*$ \\
\hline
Human {\em Homo Sapiens} \cite{AnimaK} & 442,543,570 & 467 \\
Butterfly {\em Lycaena Heterona}\cite{AnimaK} & 360,437,500,568 & 457 \\
Honeybee {\em Apis Mellifera} \cite{AnimaK} & 340,430,535 & 395 \\
Gecko {\em Gecko gecko}\cite{AnimaK} & 364,467, 521 & 409  \\
Dog {\em Canis Familiaris}\cite{neitz1989color} & 429,555 & 473 \\
Gerbil {\em Meriones unguiculatus}\cite{AnimaK} & 365,505& 413 \\
Domestic Chicken {\em Gallus gallus domesticus} \cite{hart2007avian}& 419,455,508,570 & 431  \\
\hline
\end{tabular}
\caption[]{Different species, their cone responses and the dominant wavelength in the (approximately) monochroamtic metamer to skylight.  }
\label{results-table}
\end{table}

\begin{figure}
\begin{tabular}{cccc}
human cones & human metamer & butterfly cones & butterfly metamer\\
\includegraphics[width=0.2\linewidth]{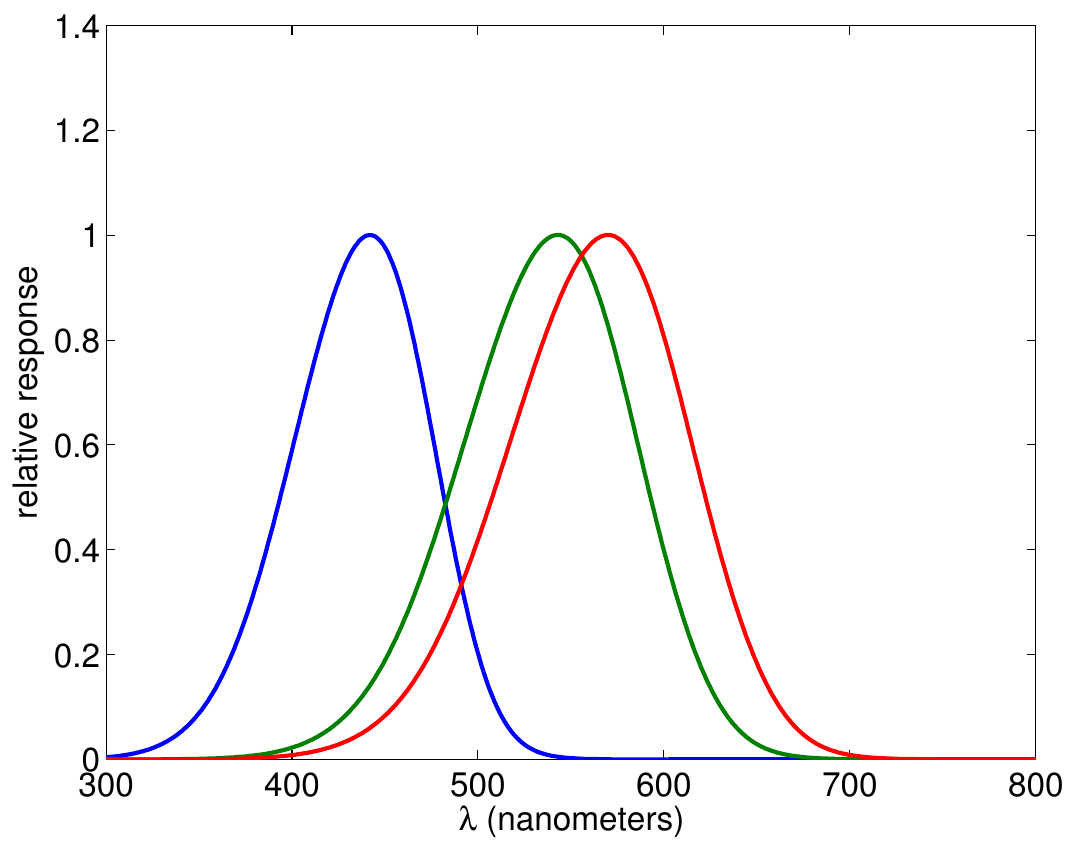} &
\includegraphics[width=0.2\linewidth]{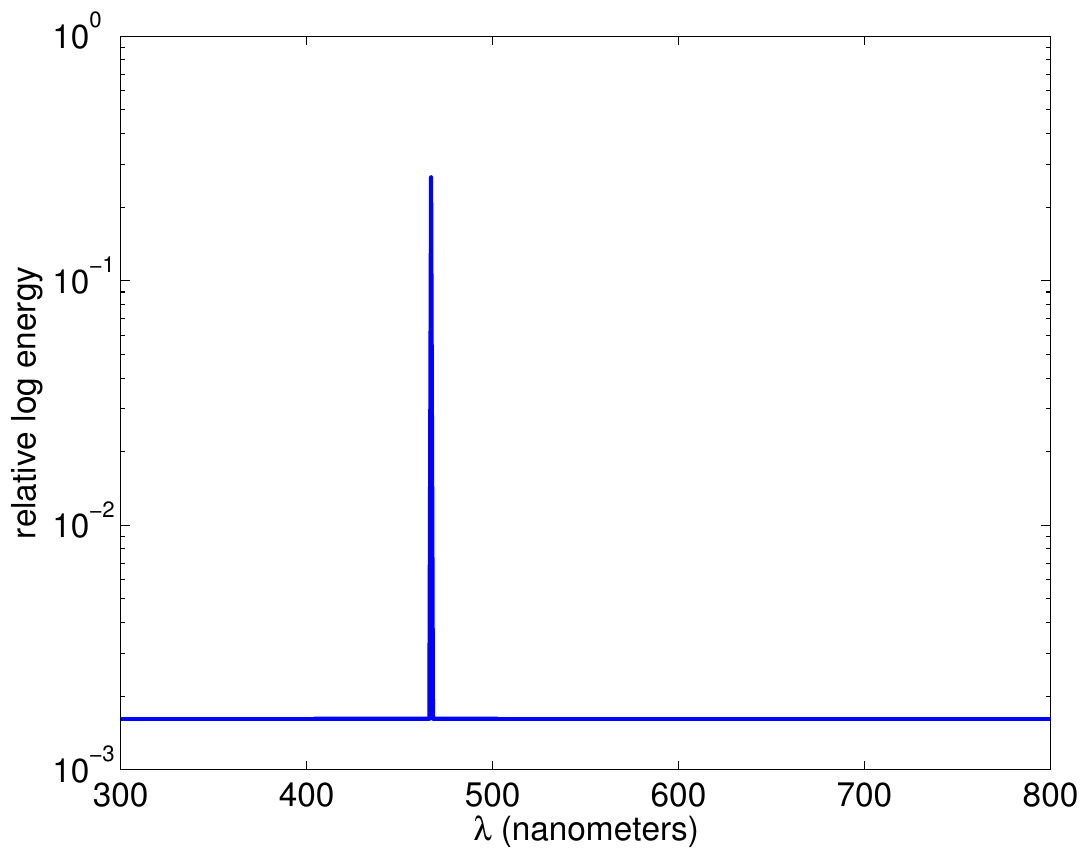} &
\includegraphics[width=0.2\linewidth]{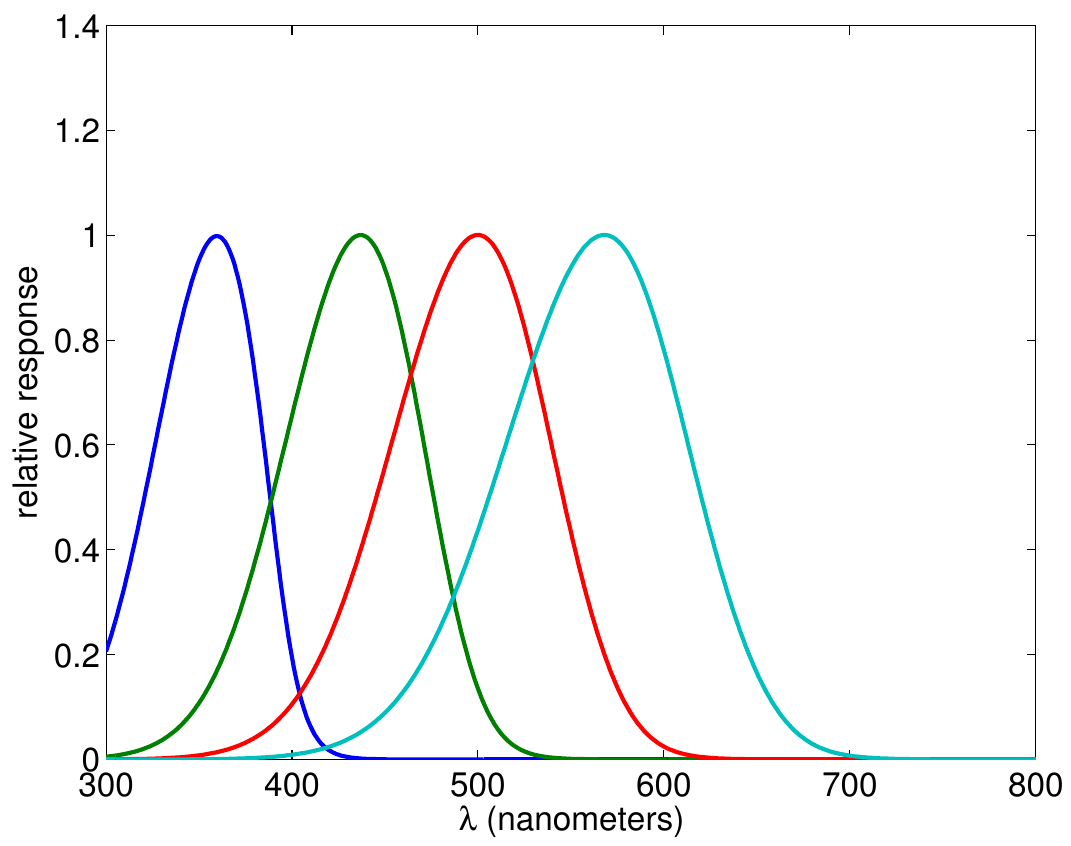} &
\includegraphics[width=0.2\linewidth]{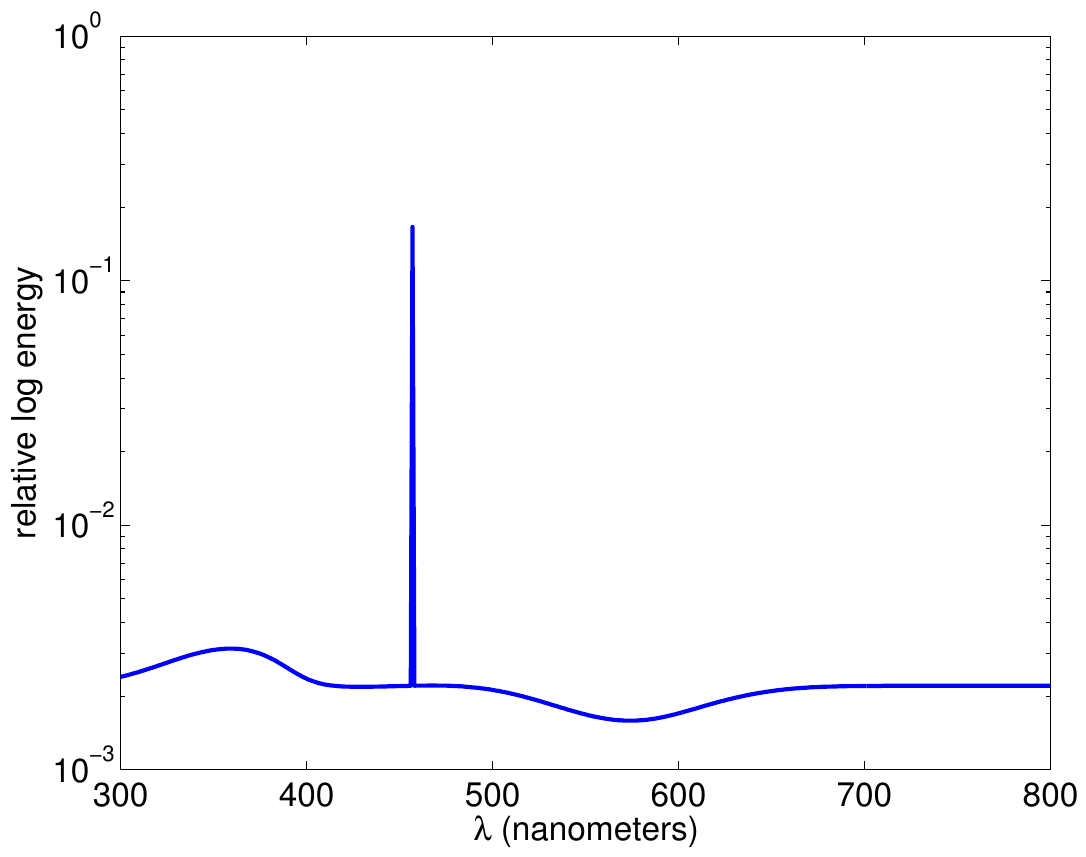} \\
honeybee cones & honeybee metamer & gecko cones & gecko metamer\\
\includegraphics[width=0.2\linewidth]{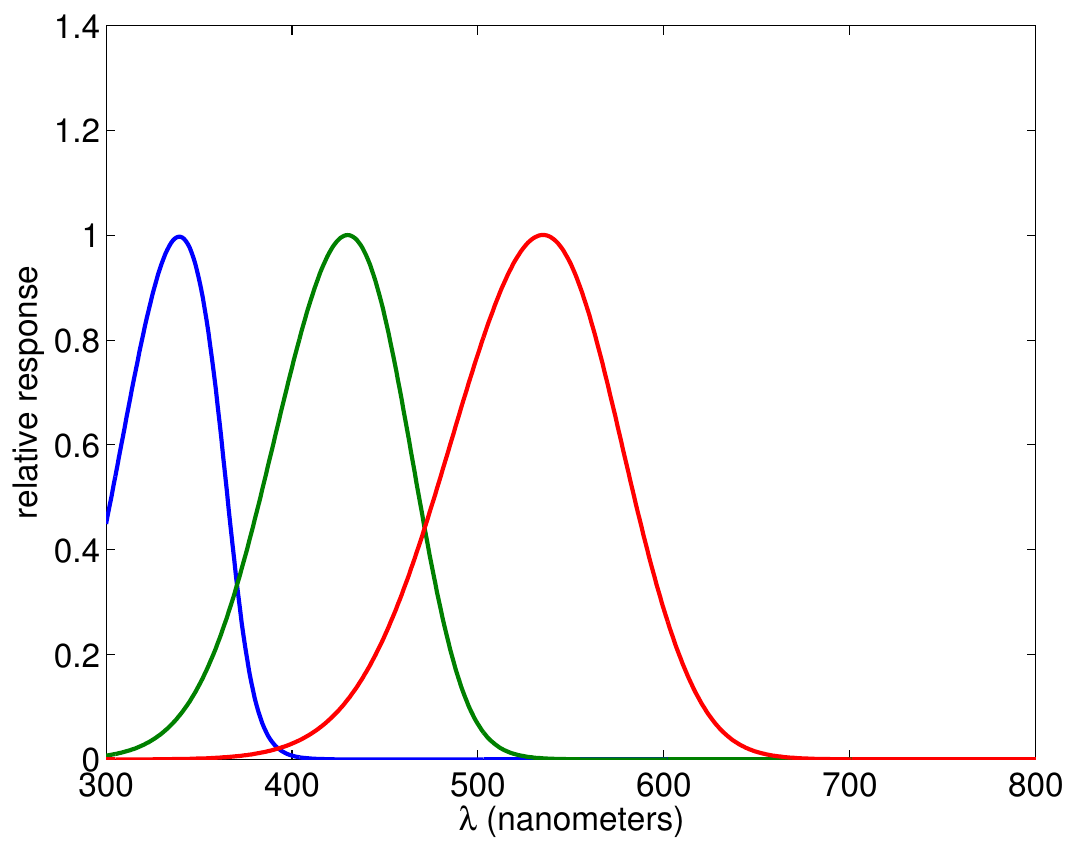} &
\includegraphics[width=0.2\linewidth]{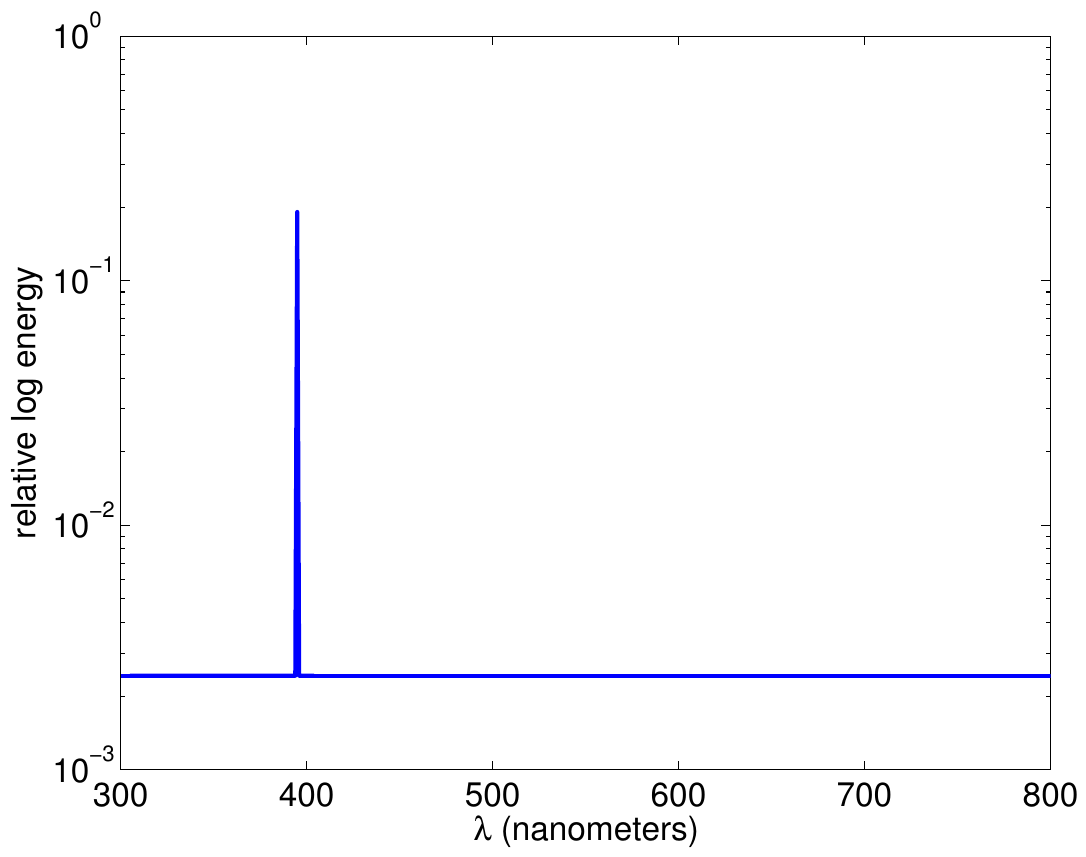} &
\includegraphics[width=0.2\linewidth]{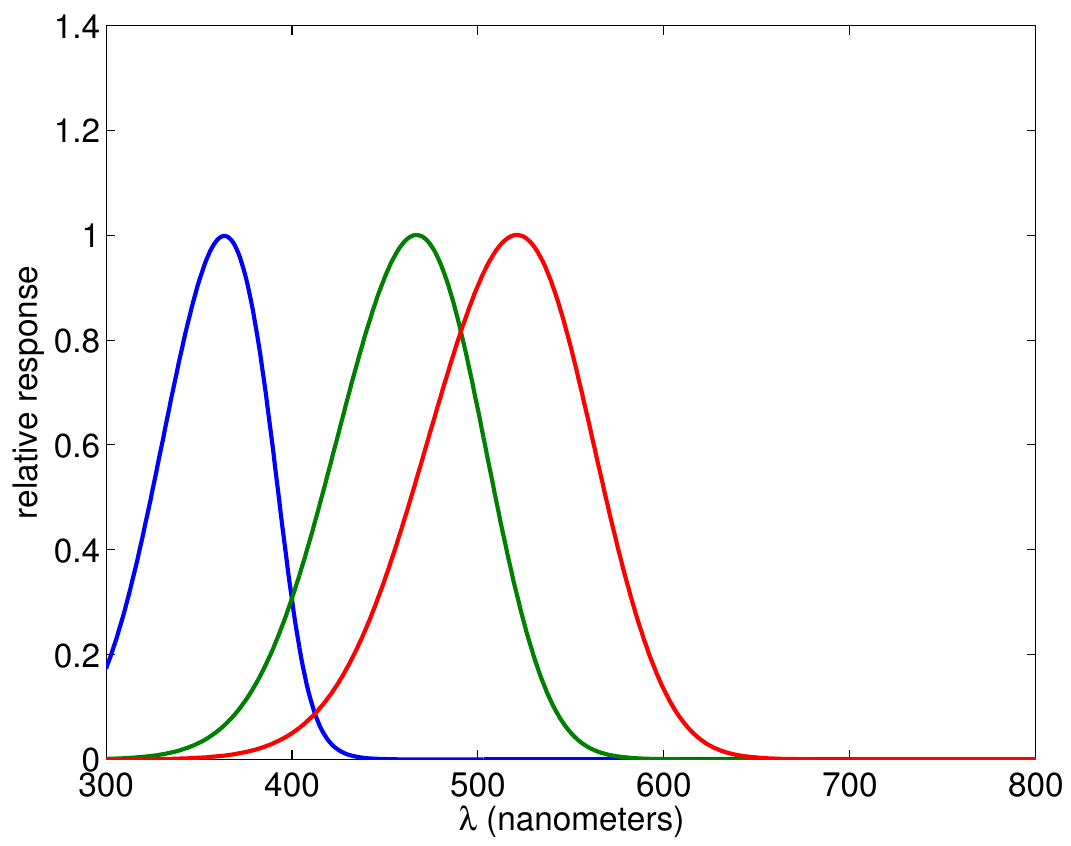} &
\includegraphics[width=0.2\linewidth]{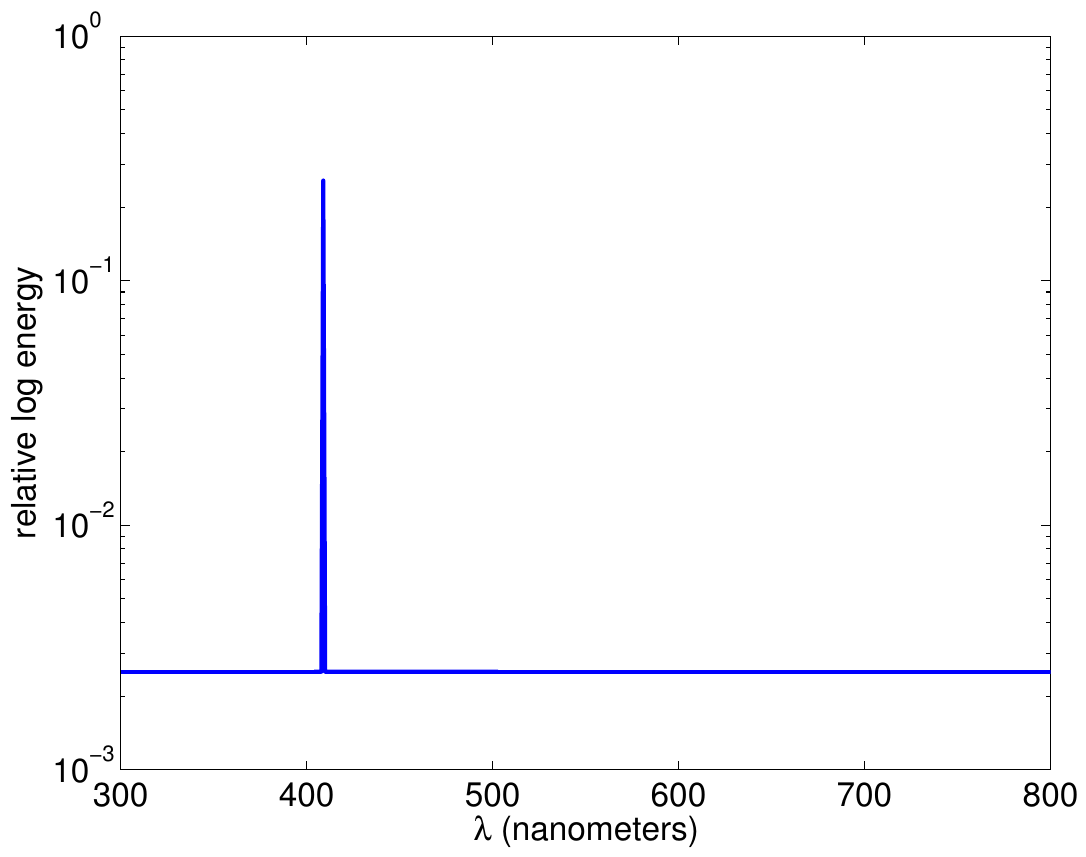} \\
dog cones & dog metamer & gerbil cones & gerbil metamer\\
\includegraphics[width=0.2\linewidth]{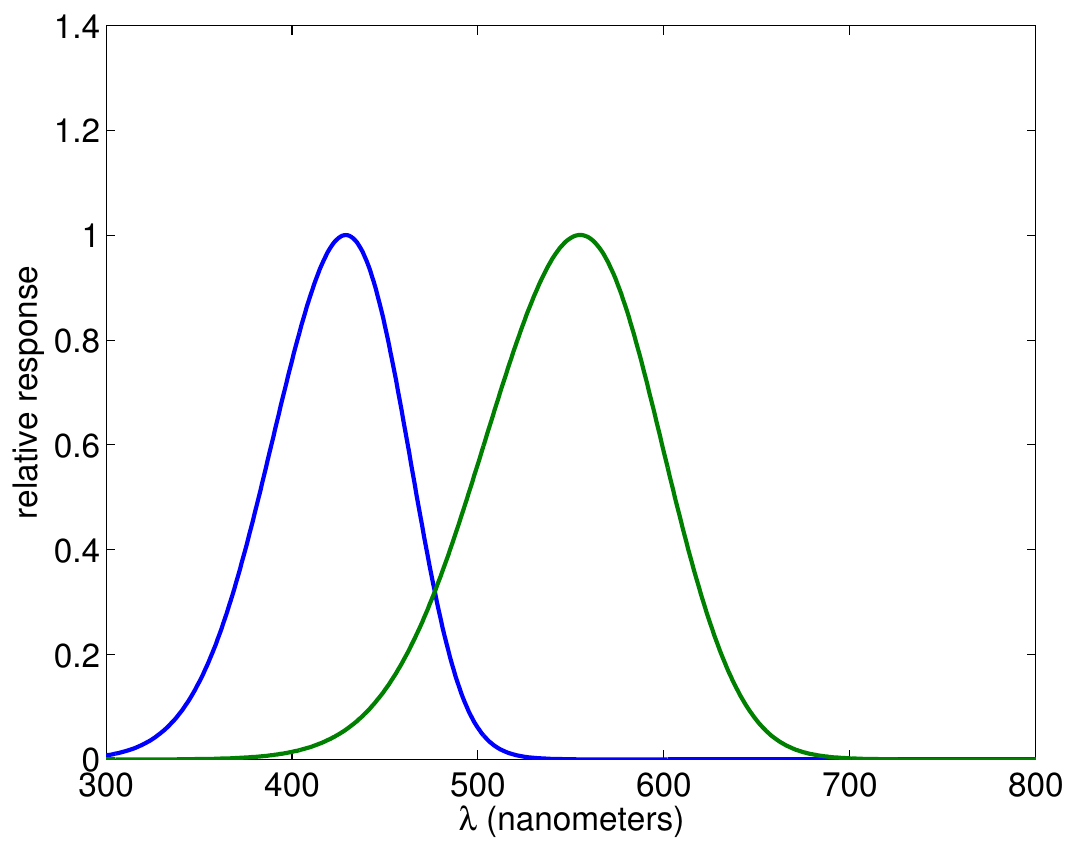} &
\includegraphics[width=0.2\linewidth]{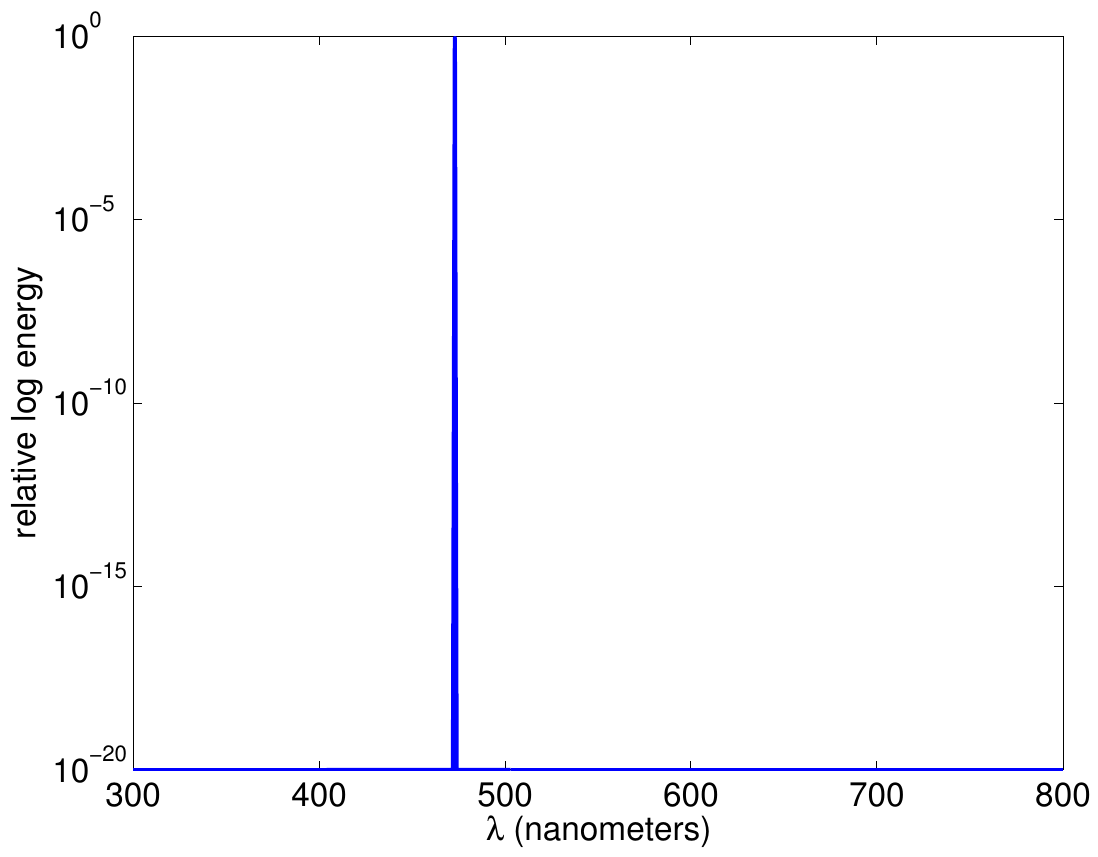} &
\includegraphics[width=0.2\linewidth]{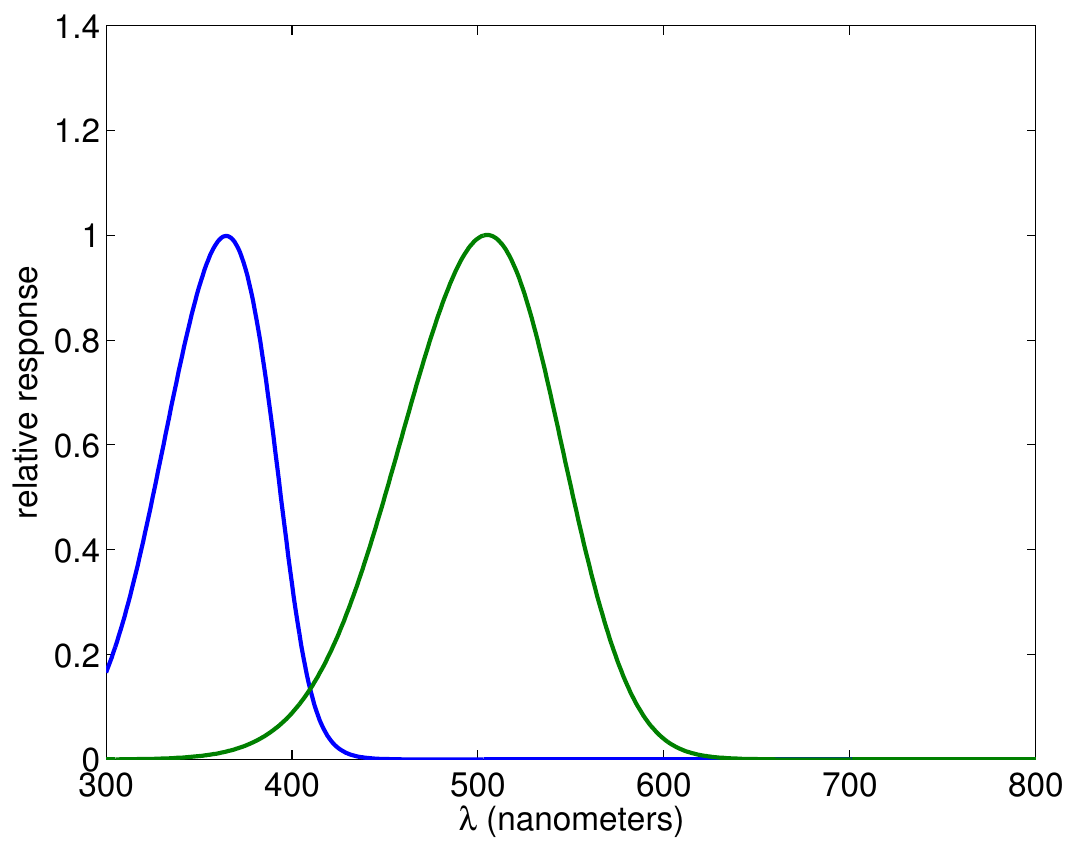} &
\includegraphics[width=0.2\linewidth]{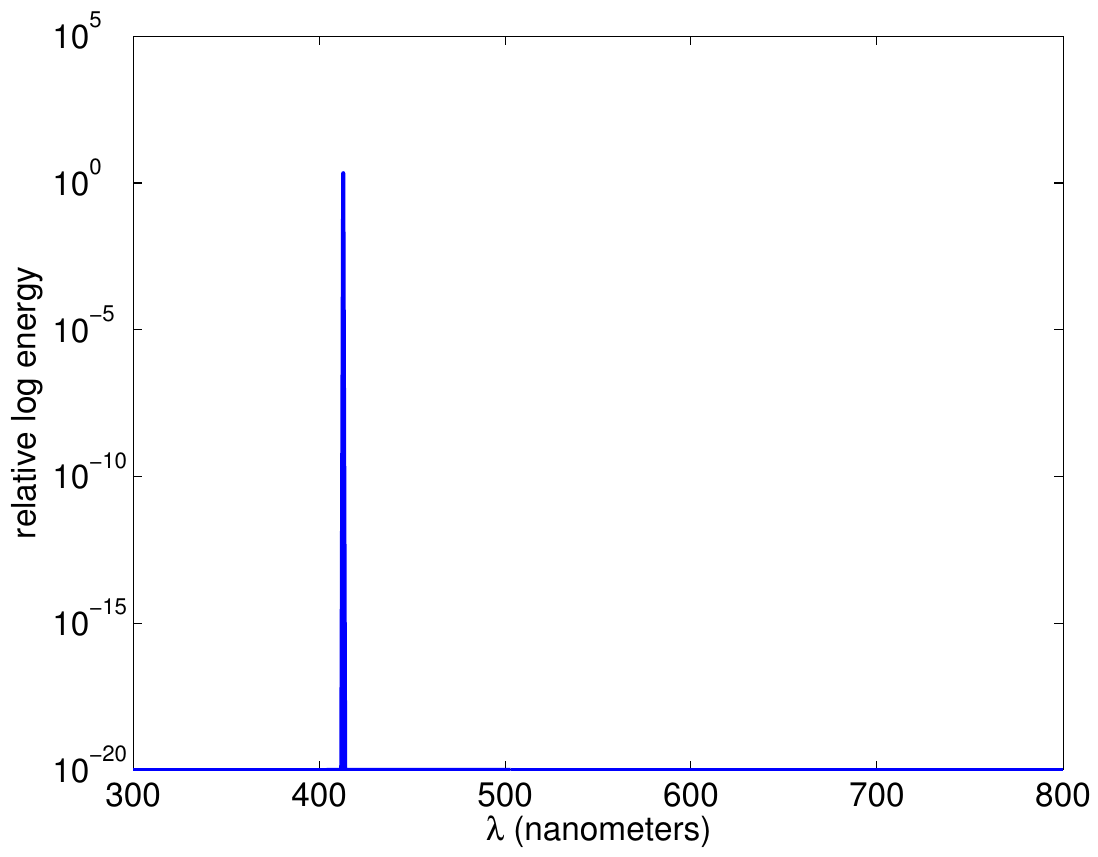} \\
chicken cones & chicken metamer & & \\
\includegraphics[width=0.2\linewidth]{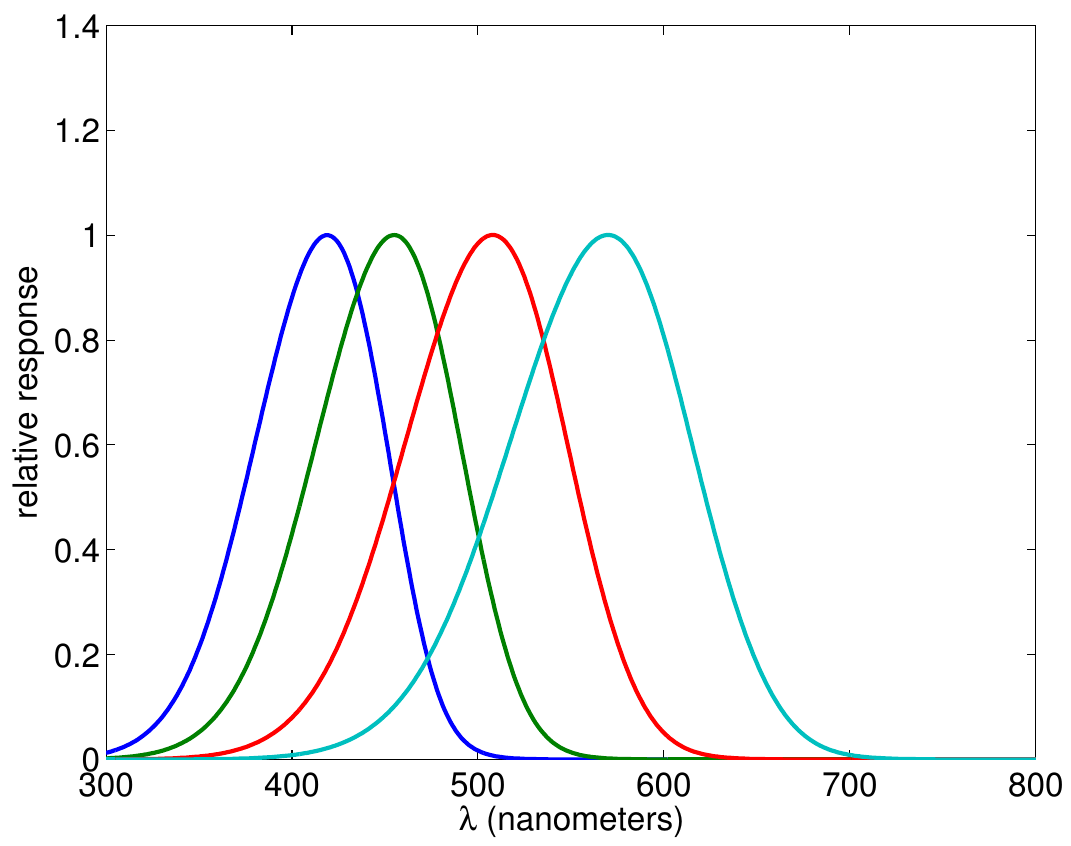} &
\includegraphics[width=0.2\linewidth]{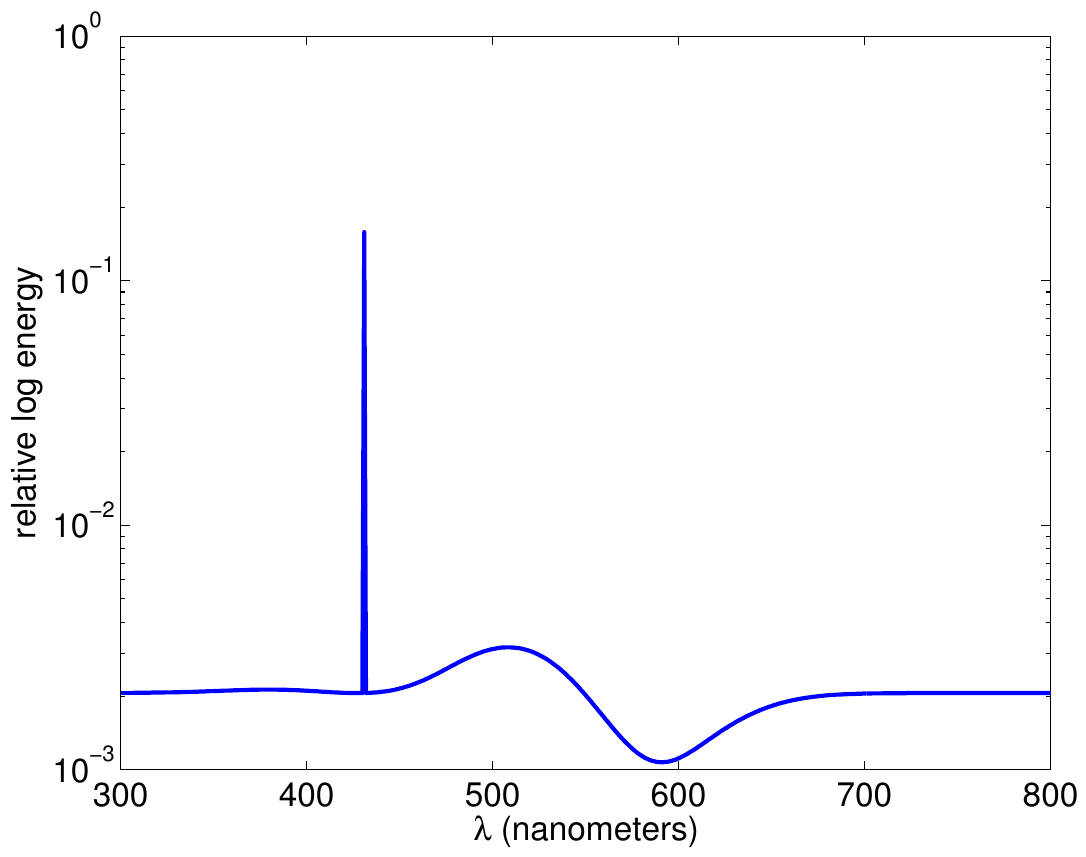} &
\end{tabular}
\caption[]{Cone spectral sensitivities and metamers to skylight for different species.}
\label{main-results-fig}
\end{figure}

Results are shown in table~\ref{results-table} and in figure~\ref{main-results-fig}. The figure shows the spectral response functions and the exact metamers of skylight for these response function. 

\begin{figure}
\begin{tabular}{cc}
\includegraphics[width=0.49\linewidth]{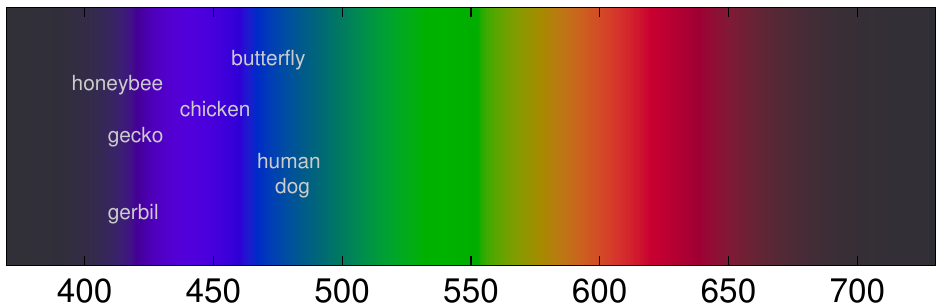} &
\includegraphics[width=0.49\linewidth]{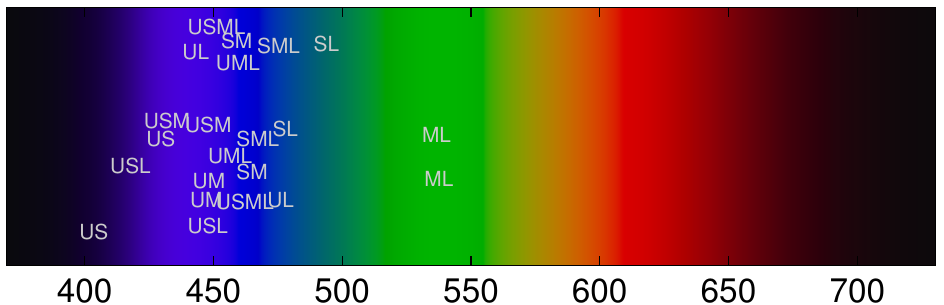} \\
a & b 
\end{tabular}
\caption[]{ {\bf a.}  A summary of our experiments with seven species. {\bf b.} A simulation of a range of possible visual systems with a simplified model of receptor evolution.}
\label{summary-fig}
\end{figure}

 As can be seen, for dichromats and trichromats we can always find an exact metamer to skylight that is either monochromatic or a mixture of white light and monochromatic light. For tetrachromats the exact metamer is dominated by white light and monochromatic light but also includes a residual signal. Regardless of the number of color receptors the wavelength of the monochromatic component of the metamer to skylight can vary anywhere between 400 ('violet') and 480 ('blue') nanometers.

Note that the human metamer to skylight  calculated in figure~\ref{summary-fig} gives a $\lambda^*$ value of $467nm$ while that in figure~\ref{human-metamer} gives a $\lambda^*$ value of $479nm$. This is because the simulation in figure~\ref{human-metamer} uses the measured human spectral sensitivity functions while the simulation in figure~\ref{summary-fig} uses the visual template and the $\lambda_{\max}$ values. This difference of only $12nm$ illustrates the robustness of our method to perturbations in the spectral response profiles.

% eye citation Eye (1998) 12, 541-547 Bowmaker. Evolution of Colour Vision in Vertebrates.

Finally, in our last experiment we wished to explore the range of
possible biological visual systems. We used a simplified evolutionary
model following~\cite{eye1998} whereby each visual system has at most
four color receptors: (1) UVS cones with peak sensitivities in the
range (350,430) (2) SWS cones with peak sensitivities in the range  (430,470) (3) MWS cones with peak sensitivities in the range  (480,530) and (4) LWS cones with peak sensitivities in the range (500,570). To create a synthetic visual system, we chose the peak
sensitivity of each type of cone from a uniform distribution over the
possible values and also performed simulated ``knock-out'': we removed
one or more of the receptors. In total we created 26 synthetic visual
systems which included dichromats, trichromats and tetrachromats. For each
such synthetic visual system we used our algorithm to calculate the
dominant wavelength of the metamers to skylight. Results are shown in figure~\ref{summary-fig}b. For each synthetic visual system we plot the dominant wavelength and label it using the receptors that were not ``knocked out'': e.g. the system labeled ``USL'' had its MWS cone knocked out and was left with the U, S and L cones. The ``color of the sky'' in these synthetic systems can range anywhere between violet and green. 
  
\section*{Discussion}

The question ``what color is the sky for a nonhuman'' may seem strange
for two reasons. First, if we believe that color perception equals
color naming then we would not expect any other species other than
humans to have a well defined ``color'' of the sky. Second, if we
believe the color of the sky is primarily determined by the physics of
skylight then we would expect all species to perceive the same
color. In this paper we have shown that both expectations are
incorrect. We have shown that it is possible to define a ``color'' of
the sky without resort to color naming but rather by constructing
monochromatic metamers. Second, we have shown that the wavelength of
the monochromatic metamer depends greatly on the form of of the
photoreceptors: the same physical skylight can be metameric to very
different monochromatic lights depending on the receptors.

% You may title this section "Methods" or "Models". 
% "Models" is not a valid title for PLoS ONE authors. However, PLoS ONE
% authors may use "Analysis" 

% Do NOT remove this, even if you are not including acknowledgments
\section*{Acknowledgments}

%\section*{References}
% The bibtex filename
\bibliography{template}

\section*{Figure Legends}
%\begin{figure}[!ht]
%\begin{center}
%%\includegraphics[width=4in]{figure_name.2.eps}
%\end{center}
%\caption{
%{\bf Bold the first sentence.}  Rest of figure 2  caption.  Caption 
%should be left justified, as specified by the options to the caption 
%package.
%}
%\label{Figure_label}
%\end{figure}

\section*{Tables}
%\begin{table}[!ht]
%\caption{
%\bf{Table title}}
%\begin{tabular}{|c|c|c|}
%table information
%\end{tabular}
%\begin{flushleft}Table caption
%\end{flushleft}
%\label{tab:label}
% \end{table}

\end{document}